\documentclass[12pt]{article}

\usepackage{arxiv}

\usepackage[utf8]{inputenc} 
\usepackage[T1]{fontenc}    
\usepackage{hyperref}       
\usepackage{url}            
\usepackage{booktabs}       
\usepackage{amsfonts}       
\usepackage{nicefrac}       
\usepackage{microtype}      
\usepackage{lipsum}
\usepackage{graphicx}
\usepackage{amsmath}
\usepackage{amsthm}
\usepackage[all]{xy}
\usepackage{array}
\usepackage[toc,page]{appendix}
\usepackage{tabularx}
\usepackage[affil-it]{authblk}

\theoremstyle{definition}
\newtheorem{definition}{Definition}

\newcommand{\pa}[2]{\frac{\partial #1}{\partial #2}}

\title{A Relational Macrostate Theory Guides Artificial Intelligence to Learn Macro and Design Micro}

\author[1,2]{Yanbo Zhang}
\author[1, 2, 3 *]{Sara Imari Walker}

\affil[1]{School of Earth and Space Exploration, Arizona State University, Tempe AZ USA}
\affil[2]{Beyond Center for Fundamental Concepts in Science, Arizona State University, Tempe AZ USA}
\affil[3]{Santa Fe Institute, Santa Fe, NM USA}
\affil[*]{author for correspondence: \texttt{sara.i.walker@asu.edu}}

\begin{document}
\maketitle

\begin{abstract}
The high-dimesionality, non-linearity and emergent properties of complex systems pose a challenge to identifying general laws in the same manner that has been so successful in simpler physical systems. In Anderson's seminal work on why ``more is different'' he pointed to how emergent, macroscale patterns break symmetries of the underlying microscale laws. Yet, less recognized is that these large-scale, emergent patterns must also retain some symmetries of the microscale rules. Here we introduce a new, relational macrostate theory (RMT) that defines macrostates in terms of symmetries between two mutually predictive observations, and develop a machine learning architecture, MacroNet, that identifies macrostates. Using this framework, we show how macrostates can be identifed across systems ranging in complexity from the simplicity of the simple harmonic oscillator to the much more complex spatial patterning characteristic of Turing instabilities. Furthermore, we show how our framework can be used for the inverse design of microstates consistent with a given macroscopic property -- in Turing patterns this allows us to design underlying rule with a given specification of spatial patterning, and to identify which rule parameters most control these patterns. By demonstrating a general theory for how macroscopic properties emerge from conservation of symmetries in the mapping between observations, we provide a machine learning framework that allows a unified approach to identifying macrostates in systems from the simple to complex, and allows the design of new examples consistent with a given macroscopic property.  

\end{abstract}

\keywords{Emergence \and macrostates \and physics \and complex systems \and machine learning}

\section{Introduction}

Among the most important concepts in physics is that of symmetry, and
how symmetry-breaking at the microscale can give rise to macroscale
behaviors. This deep connection was made clearest in the work of Noether~\cite{noether1918invariante}
, where she showed that for differentiable systems with
conservative forces, every symmetry comes with a corresponding
conversation law that describes macroscale behavior. An example is how
time translation symmetry gives rise to the conservation of energy:
simple harmonic oscillators conserve energy because, in the absence of
friction, you will observe the same oscillation if starting a clock at
the first cycle as at the thousandth -- the behavior is time invariant.
Thus, Noether's theorem provided a means to relate laws -- namely,
regularities that are conserved (e.g., energy conservation) -- to
symmetries in the underlying physical system (e.g., time). Physics has
been incredibly successful at discovering laws in this manner. However,
so far, finding similar `law-like' behaviors for complex systems, such
as biological and technological ones, has proved much more challenging
because of their high-dimensionality, non-linear behavior, and emergent
properties. Yet, the very concept of emergence provides a clue that such
regularities should exist, even for complex systems. In Anderson's
seminal work on why ``more is different''~\cite{anderson1972more}, he pointed to
how symmetry-breaking also plays a prominent role in emergence:
macroscale behaviors do not necessarily share all the same symmetries as
the microscale laws or rules that give rise to them. While some of the
symmetries are clearly lost, this also leaves open the possibility
large-scale patterns that emerge will still retain other symmetries of
the microscale rules. In addition to the rule-behavior mapping, there
are other mappings unique to complex systems such as genotype-phenotype
maps, text-image maps, etc., where symmetries may lead to conserved
properties. The challenge to identifying general laws for complex
systems then reduces to identifying which symmetries are preserved
during the mapping -- in general this is challenging because of their
high dimensionality, suggesting that machine learning might be an
approach that can aid in identifying conservation laws in these systems,
if we can identify macrostates and the symmetries they retain from the
microscale.

\begin{figure}
    \centering
    \includegraphics[width=6.5in]{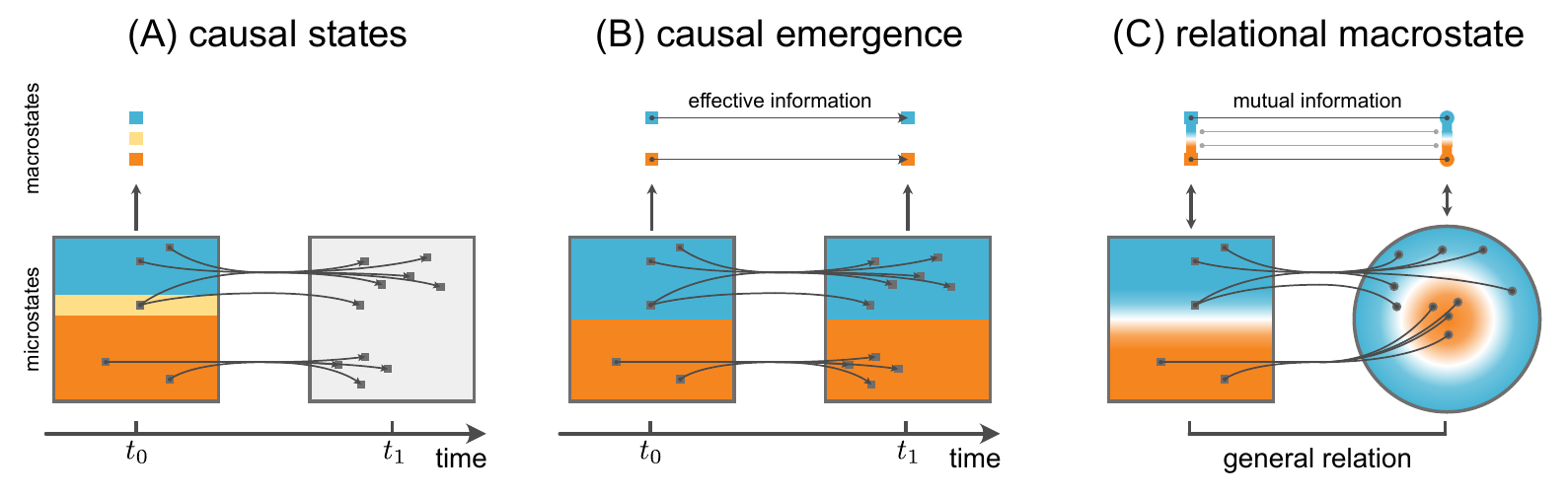}
    \caption{Comparison between causal state theory, causal emergence theory
and relational macrostate theories \textbf{(A)} In causal state theory,
two microstates are equivalent (belong to the same macrostate) if their
future microstate distributions are the same. \textbf{(B)} Causal
emergence theory identifies macrostates where the past-future mappings
are deterministic and non-degenerate at the macro scale, such that the
macrostates can be distinguished from one another in the past
(non-degeneracy) and the future (determinacy). Both causal state theory
and causal emergence theory define macrostates in terms of temporal
relations within a system of interest, as denoted by the square shape
underlying the mapping. \textbf{(C)} In the relational macrostate theory
we propose here, two microstates are equivalent if they relate to the
same microstate distributions, which can be generalized to any type of
relation, including past-future, rule-pattern, genotype-phenotype, etc.
-- the square and disk shapes denote that this is sufficiently general
to apply to maps that exist across different spaces.}
    \label{fig:compare}
\end{figure}

There have been several efforts focused on identifying macrostates
associated with the emergent regularities found in complex systems~\cite{chen2021discovering, kipf2019contrastive, sun2021eigen}. Notably,
Shalizi and Moore proposed causal state theory~\cite{shalizi2003macrostate},
which defines macrostates based on the \emph{relations between
microstates}. Here, two microstates are equivalent (belong to the same
macrostate) if the future microstates distributions are the same (Figure~\ref{fig:compare}A). Thus, the conserved symmetry is one pertaining to the prediction of
future states. This, however, can exclude some well-defined macrostates
in physics. For example, given a simple harmonic oscillator, two
distinct microstates $u_{1} = (p_{1},x_{1})$ and
$u_{2} = (p_{2},x_{2})$, where $p_{1}$ and $p_{2}$ are two
observations of momentum and $x_{1}$ and $x_{2}$ there corresponding
position, can have the same energy macrostate. However, their future
microstate distributions will be different if $u_{1}$ and $u_{2}$
are not close to each other, say if, $u_{1} = - u_{2}$. The conserved
symmetry of Shalizi and Moore is therefore violated because this system
does not retain predictability of future microstate distributions at the
macroscale (because the macrostate of energy is related to the time
translation symmetry, not the symmetry associated to predictability of
future states).

If a proposed theory to define macrostates is not sufficiently general
to include simple physical examples like the harmonic oscillator, it is
unlikely to apply universally to complex systems. Indeed, Shalizi and
Moore were not looking for a general theory of macrostates, but instead
focused on the specific property of predictability of complex systems.
Another approach was more recently proposed in causal emergence theory~\cite{hoel2017map, chvykov2020causal, comolatti2022causal}, which likewise has a specific
goal in mind -- to describe causal relations at the macroscale. Here,
instead of using the properties of microstates, macrostates are defined
based on the \emph{relations between macrostates} by maximizing
effective information at the macroscale. Effective information is the
mutual information between two variables, under intervention to set one
of them to maximum entropy (e.g., a uniform distribution over
macrostates). Causal emergence occurs when the past and future of
different macrostates are distinguishable (Figure~\ref{fig:compare}B). Thus, the symmetry
of distinguishing past and future leads to a conservation of
distinguishability of macrostates.

Both causal state theory and causal emergence theory define macrostates
in terms of \emph{temporal} relations between past and future. However,
not all regularities we might want to associate to laws involve time.
For instance, to get the macrostates of mass, force, and acceleration,
physicists of past generations needed to study the relations between two
objects rather than between points in time (past and future). This
suggests that to develop a general theory of macrostates, these must be
defined based on general relations between two observations (Figure~\ref{fig:compare}C)
that retain some of the symmetries of the underlying sets of
observations -- those observations could be objects, or points in time,
as physics has already treated. But they can also be any other
observation we can make with a measuring device, including more
``complex'' examples like genotype-phenotype maps, or word co-occurrence
in language, or rule-behavior mapping necessary to describe patterning
in biological form. As we will show, the theory of macrostates we
propose is sufficiently general to extend to the symmetry of the
microscale level rules (or laws), which allow us to identify sets of
microscale rules that yield a given emergent, macroscale behavior~\cite{anderson1972more}.

When studying the history of the laws of physics, it is important to
identify why the most successful laws have worked so well. Newton's laws
of motion work because there is a macroscale property called mass, which
quantifies the amount of matter in each object, that reduces the
description of the motion of high dimensional objects to a single
measurable scalar quantity (mass) and its translation in $x,y,z$
coordinates. For complex systems it is not so obvious what the necessary
dimensionality reduction will be that allows identifying law-like
behavior, and it may vary from system to system. Of note, Newton's laws
cannot be developed in a world where mass can only be defined and
measured in a few countable objects and is undefined or unmeasurable in
others. In the current work, we show how artificial neural networks,
themselves a complex system, can break the barrier of complexity~\cite{jumper2021highly, pathak2018model, seif2021machine} to identify
macrostates based on symmetries in complex systems. Existing machine
learning methods such as contrastive learning~\cite{chen2021exploring},
contrastive predictive coding~\cite{oord2018representation}, and
word2vec~\cite{mikolov2013efficient} have applied similar ideas to find lower
dimensional representations for microstates by relations. However, these
contrastive methods either require large numbers of negative samples
that increases the cost of training, or only learning embeddings instead
of functional mappings. Moreover, these methods are only useful for
downstream tasks, which use the embedding trained by contrastive
learning. Although we describe things at the macroscale, the world still
runs on microscale features. This means we not only need to map
microstates to macrostates, but that we need to provide an inverse path
that samples microstate from a given macrostate. By developing the
macrostate theory on general relations, and introducing invertibility,
we propose a machine learning architecture, MacroNet, that can learn
macrostates and design microstates.

In fact, a key feature of learning is demonstrating use cases of the
knowledge learned. Therefore, to demonstrate how MacroNet is indeed
learning the macrostates across of examples of simple physical systems
and complex systems, we also use it to design new examples. There has
been a flurry of recent work by scientists attempting to engineer AI
scientists, and in particular AI physicists, that can learn the laws of
nature from data with minimal supervision. Examples include: AI Feynman,
which learns symbolic expressions~\cite{udrescu2020aifeynman}; 
AI Poincare that can learn conservation laws~\cite{liu2021machine};
and Sir Isaac, an inference algorithm that can learn dynamical
laws~\cite{daniels2015automated}. Yet, science as done by scientists
goes further than solely extracting laws from data -- humans also
implement that understanding in the real world. For example, in the case
of Newton's laws of motion, our knowledge of them has allowed us to
engineer a range of systems, such as the design of airbags, racecars,
airplanes, helicopters and even optimization of athlete performance,
etc. Thus, we view the next advancement beyond artificial intelligence
that can learn the rules by which data behave is AI that can also use
that knowledge to design new examples of systems that will behave by the
rules identified. A critical aspect of designing new examples of systems
is identifying macrostate variables that reduce high-dimensional data to
a few variables that capture the salient features. The invertibility of
MacroNet not only allows the design of microstates sampled from an
identified macrostate, but also provides a low-cost way to replace
negative samplings in contrastive learning.

In what follows, we first introduce our mathematical framework for
defining macrostates in terms of relations defined by symmetries in the
data. Then, we propose a machine learning framework to find macrostates
under the definition. For experiments, we first demonstrate the workflow
of the framework by implementing it for linear dynamical systems. They
are simple enough to demonstrate key concepts, but also exhibit rich
behaviors. Then, we introduce the simple harmonic oscillator as a
special case where macrostates are defined based on temporal relations,
which demonstrates how our framework can extract familiar invariant
macrostates (conserved properties associated to symmetries) from
physics, such as energy. Finally, we turn to a real complex system, the
macroscale Turing patterns that arise in diffusion reaction systems. We
show how machine learning finds the macrostates associated with the
emergent patterning in these systems, and then how this can be used to
design microstates consistent with a target macroscale pattern.

\section{Theory and Method}

\subsection{The relational macrostate theory that generalizable to the study of complex systems using neural networks}

\begin{figure}
    \centering
    \includegraphics[width=6in]{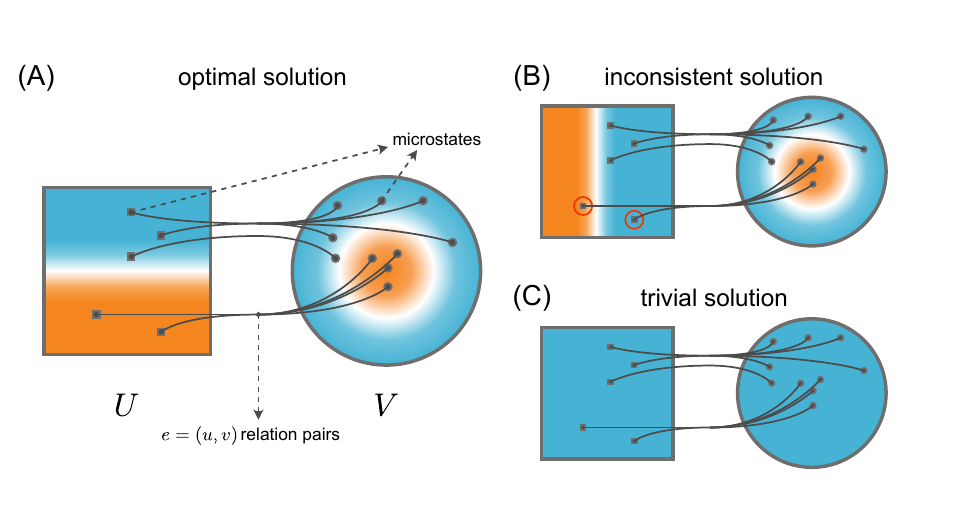}
    \caption{Macrostates are defined by symmetries that define relations
between ensembles of microstates. The rectangle and disks represent the
space of microstates. And the points and links represent the observed
microstate pairs $( u_{i},v_{i} )$. The color illustrates
the macrostates that represent the underlying symmetries in the
relations between microstates. \textbf{(A)} shows an optimal solution. \textbf{(B)} shows
a trivial but legal solution, which coarse grains all microstates to the
same macrostate. This kind of coarse graining is not informative since
the mutual information of macrostates is zero. \textbf{(C)} shows an inconsistent
solution.}
    \label{fig:theory}
\end{figure}

By definition, a macrostate is an ensemble corresponding to an
equivalence class of microstates. Given a mapping $\varphi_{u}$ that
maps microstates to macrostates, two microstates $u$ and $u'$
belong to the same equivalence class if
$\varphi_{u}(u) = \varphi_{u}( u' )$, that is if the
microstates have the same behavior (macrostate) under the operation of
the map. In this way, macrostates are also the parameters to describe
distributions of microstates. This feature is a key reason why machine
learning may be an optimal way to identify macrostates, particularly in
cases of many-to-many mappings such as those that occur in rule-behavior
maps, or under prediction with noise, both of which are characteristic
of complex systems.

Here, we implement a formalism based on using relations arising due to
symmetries to define macrostates. Consider two microstates $u\in U$ and
$v\in V$ as two random variables. Their \emph{micro-to-micro relation} can
be mathematically represented as a joint distribution $P(u,v)$. The
$u$ and $v$ can be mapped to macrostates $\alpha$ and $\beta$
respectively by $\varphi_{u}$ and $\varphi_{v}$. So, we can also
define \emph{micro-to-macro relation} by the joint distribution
$P(\alpha,v)$ and $P(u,\beta)$. For a given microstate $u_{i}$ (or
$v_{i}$), its micro-to-macro relation can be represented as a
conditional distribution $\Pr(\beta|u_{i})$ (or
$\Pr(\alpha|v_{i})$). Then, we can define macrostates in the most
(relational) general case as:

\begin{definition}
    Two pairs of microstates
$u_{i}$ and $u_{j}$ (and $v_i$ and
$v_j$) belong to the same macrostate if and only if
they have the same \emph{micro-to-macro relation}:
    \begin{align}
        u_i\sim u_j &\iff \Pr(\beta|u_i)=\Pr(\beta|u_j) \text{ and}\\
        v_i\sim v_j &\iff \Pr(\alpha|v_i)=\Pr(\alpha|v_j)
    \end{align}
    \label{def:macro}
\end{definition}

Note, this defines an equivalence class of symmetries where
$u_{i} \sim u_{j}$ and $v_{i} \sim v_{j}$ (where $\sim$
indicates ``is equivalent to'' under the symmetry operation). Thus, as in
Noether's theorem (and in Anderson's formalization of emergence) we see
that the definition of a macrostate entails simultaneously defining a
class of symmetry operations, although here our definition is
sufficiently general that the system of interest need not necessarily be
continuously differentiable (as in the case of Noether's theorem).

The definition can be approached by solving
$\varphi_{u}(u) = \varphi_{v}(v)$ (see SI). This equation will be part
of the loss function in the specified machine learning task of MacroNet.
Since the macrostate of $U$ is defined by the macrostate of $V$, and
vice versa, the solutions are not computed in a straightforward way, but
must be calculated in relation to one another. As such, there can exist
some inconsistent solutions. Figure \ref{fig:theory}A shows a consistent solution,
however, Figure \ref{fig:theory}B shows an inconsistent solution. The points in red
circles are classified into two macrostates, while they both have the
same micro-to-macro relation. Not all consistent solutions are useful.
If all microstates are mapped to the same macrostate, it still follows
the definition, but this is a trivial solution and not informative, see
Figure~\ref{fig:theory}C. In addition to definition~\ref{def:macro}, we therefore require an
information criterion to specify ``good macrostates''. We do so by
specifying a given dimension of macrostates, and then maximizing the
mutual information $I\left( \varphi(v);\varphi(u) \right)$ at the
macroscopic level, where $(u,v)$ is sampled from $P(u,v)$. As a
comparison, the effective information (EI) in causal emergence theory
also uses the mutual information concept, but it has notable differences
in how it is implemented beyond the fact that the theory presented here
was designed for a machine learning implementation and causal emergence
was not. In a discrete macrostate space, to quantify the causal effect,
the EI re-assigns the marginal distribution of the macrostates with a
uniform distribution. We do not make this requirement since we are not
focusing on causal relations. Moreover, in a continuous macrostate
space, the causal relation between macrostates may not make much sense
because of the large number of different macrostates.

\subsection{A self-supervised generative model for finding macrostates from observations}

\begin{figure}
    \centering
    \includegraphics[width=6.5in]{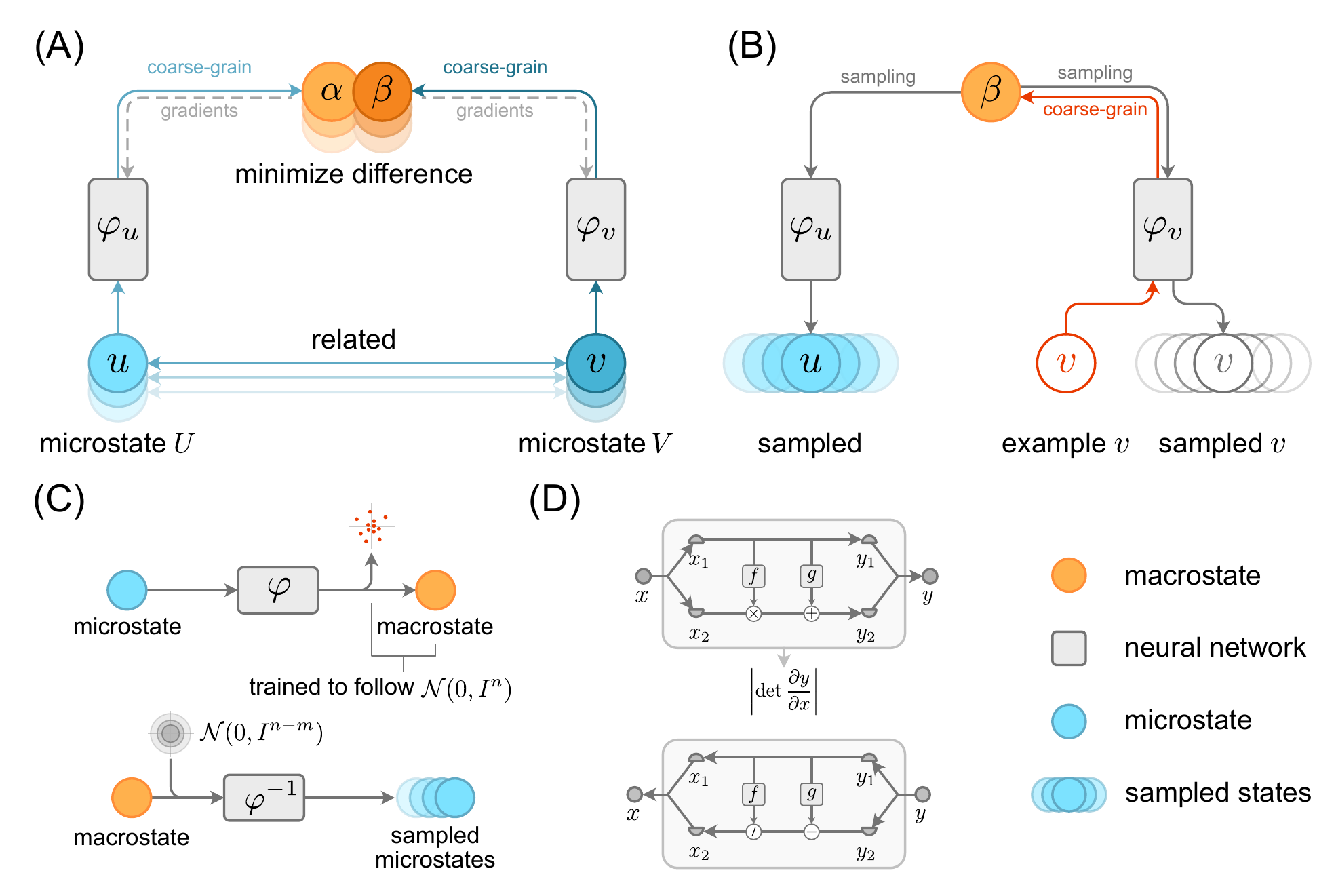}
    \caption{Neural network architecture of MacroNet. \textbf{(A)} During
the training process, the two invertible neural networks are optimized
to map two types of microstates to the same macrostates. These
microstates can correspond to past and future states of the same
dynamical system, dynamical rules or parameters and observed behavior,
or any other pair of (sets of) variables \textbf{(B)} The conditional
sampling and designing process. First, we can manually make an example
microstate of type $V$. We compute the macrostate. We can sample the
microstates in $U$ or $V$ that have this macrostate. \textbf{(C)}
When doing coarse-graining, parts of the output are abandoned to reduce
dimensionality. The abandoned variables are still trained to follow an
independent standard normal distribution. This independence makes it
easy to do conditional sampling because we can sample the abandoned
variables easily. \textbf{(D)} A typical invertible neural network is
RealNVP, which has a specially designed structure that guarantees
invertibility. The log-determinate of the Jacobian is also easy to
compute for this type of neural network.}
    \label{fig:architecture}
\end{figure}

In the above formalization, a macrostate in $U$ is defined by
macrostates in $V$ (i.e., macrostates are defined only in terms of
their relations to other macrostates). This relational definition
necessitates that we optimize the macrostate mapping iteratively to find
an optimal solution. Thus, to implement the relational macrostates
theory, we propose a self-supervised generative model for finding
macrostates from observations (Figure~\ref{fig:architecture}A).

Our definition of macrostates can be achieved by optimizing macrostates
to predict other macrostates. Here we use $\varphi_{u}$ and
$\varphi_{v}$ to represent the coarse graining performed by the neural
networks on $U$ and $V$ respectively. We have the prediction loss:

\begin{equation}
    \mathcal L_{P} = \mathbb{E}_{(u,v) \sim P(u,v)}\left| \varphi_{u}(u) - \varphi_{v}(v) \right|^{2},
\end{equation}

where $(u,v)$ are pairs of microstates sampled from the training data.
The ideal solution for $\varphi$ is
$\varphi_{u}(u) \approx_{\sigma}\varphi_{v}(v)$, meaning the
macrostate of $u$ can be predicted by the macrostate of $v$ with
error of $\sigma$, and vice versa. However, we need an additional term
to avoid trivial solutions such as a low dimensional manifold or
constant. To do this, we add a distribution loss,
$\mathcal L_{D} = \mathcal L_{D_u} + \mathcal L_{D_v}$, where:

\begin{align}
    \mathcal L_{D_u} &= \log{P_\text{normal}\left( \varphi_{u}(u) \right)} - \log\left| {\det\frac{\partial\varphi_{u}(u)}{\partial u}} \right|,\\
    \mathcal L_{D_v} &= \log{P_\text{normal}\left( \varphi_{v}(v) \right)} - \log\left| {\det\frac{\partial\varphi_{v}(v)}{\partial v}} \right|.
\end{align}

The distribution loss is be minimized when the outputs follow
independent normal distributions. We train the neural networks by
combining the two loss functions:

\begin{equation}
    \mathcal L = \mathcal L_{P} + \gamma \mathcal L_{D},
\end{equation}

where $\gamma$ is the hyperparameter balancing the two loss terms.
Combining these two terms, we can approach the mutual information
criterion. Directly computing $\mathcal L_{D}$ can be very expensive since it
requires computing the Jacobian. However, since we want to do sampling,
invertible neural networks (INNs)\footnote{We developed an invertible
  neural network python package -- INNLab, available on GitHub:
  \url{https://github.com/ELIFE-ASU/INNLab}} can help. The INNs are not
only designed to be invertible, but also designed to easily compute the
log-determinate of the Jacobian. The INNs will have the same output
dimension as the input, so we abandon part of the dimensions (see SI
S.3.c). For example, if we want to map an 8-dimensional vector to
two-dimensional macrostate, the INNs will still give an 8-dimensional
vector as a result, but we only take the first two variables as the
macrostate for training. The abandoned six variables, however, still
have been trained to follow independent normal distributions so we can
do conditional inverse sampling.

Given an example microstate $v'$, suppose we want to find other
microstates in $V$ space with the same macrostate as $v'$. We can
use $\varphi_{v}( v' )$ to compute the macrostate
$\beta$ of the example $v'$. Then, we can invert the neural
network to sample microstates $v_{s}$ that have the same macrostates.
This conditional sampling allows identifying the symmetry of macrostates
and enables the design of microstates by sampling from a given target
macrostate once the network is trained on other examples with the same
macroscale behavior (Figure~\ref{fig:architecture}B). This kind of sampling can enable the
design of complex systems: the identified macrostates are not given by
humans, but instead, computed from examples by neural networks. This
process makes it possible to design complex systems without needing to
first classify their behavior.

\section{Results}

In what follows we consider three explicit examples of the application
of MacroNet. The first is a linear dynamical system, which allows us to
demonstrate the key features of our workflow with a system that allows
easily demonstrating key concepts, via the identification of a
rotational symmetry and design of microstates consistent with this
behavior. The second example is a simple harmonic oscillator (SHO),
where we demonstrate MacroNet can identify a familiar symmetry and its
corresponding macrostate in physics -- time translation invariance and
energy -- by showing that our workflow can identify equal energy
surfaces for the SHO. The final example is Turing patterns, where we
show the utility of MacroNet in solving the inverse problem of mapping
macro-to-micro in a complex system.

\subsection{Linear dynamical systems}

\begin{figure}
    \centering
    \includegraphics[width=6.5in]{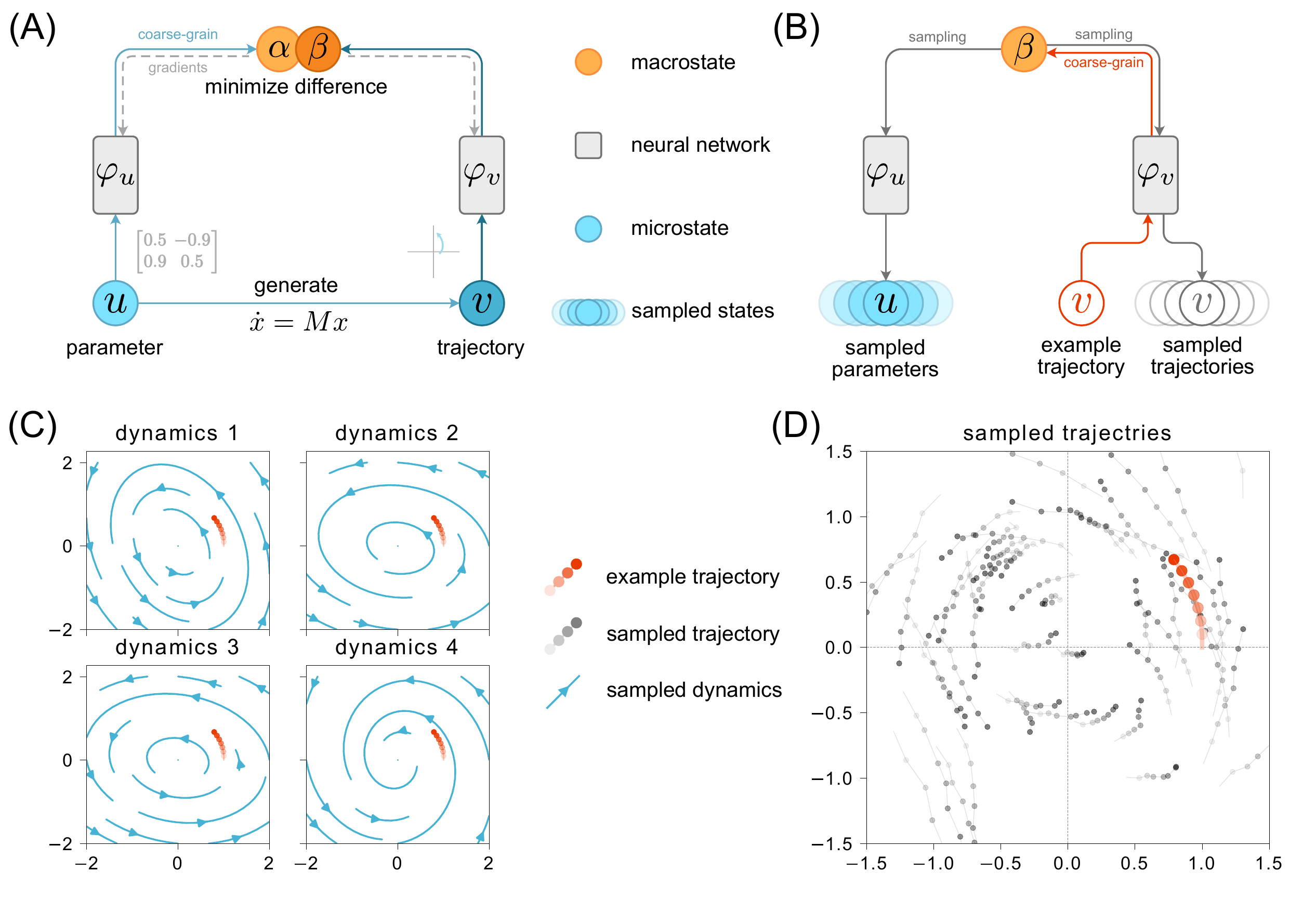}
    \caption{Training neural networks to find macrostates of linear
dynamical systems. \textbf{(A)} the related pairs are parameters and
trajectories. \textbf{(B)} by choosing an example trajectory, we can sample
microstates of parameters or trajectories with the same macrostate. \textbf{(C)}
Given the example trajectory (red dots), we can compute its macrostate.
Then, we can sample parameters that have the sample macrostate. Using
the sampled parameters, we can plot the trajectories generated by the
sampled parameter (blue lines). \textbf{(D)} Using the same macrostate, we can
sample an ensemble of trajectories that have the same macrostate.}
    \label{fig:linear}
\end{figure}

We start with an experiment analyzing a linear dynamical system because
these have many-to-many mappings. This allows us to demonstrate the
workflow of identifying macrostates based on symmetries and then
designing microstates from the identified macrostates. Here we choose a
two-dimensional linear dynamical system whose dynamics are given by

\begin{equation}
    \frac{\text d\vec x}{\text dt}=M\vec x,\label{eq:linear_dynamics}
\end{equation}

where $x$ is the independent variable, and $M$ is a $2 \times 2$
matrix that includes the parameters that specify the dynamics of the
system. Given a matrix $M$ and an initial state $x_{0}$, we can
generate a sequence of observed states by computing
$x_{t + 1} = x_{t} + Mx_{t}\delta t$. The trajectory will be
$T = \lbrack x_{1},x_{2},\ldots,x_{n}\rbrack$ in the two-dimensional
space, where $n = 8$ and $\delta t=1/n$. Here we choose $n = 8$ because it is large
enough to show the pattern of trajectories and not too large to slow the
training. In this example, the micro-to-micro relations are represented
by parameter-trajectory pairs, i.e., $(u,v) = (M,T)$. Note, in
contrast to more standard approaches to studying dynamical systems, we
are here not trying to find a macrostate by coarse graining the
trajectory of states (which would depend on some variety of time
symmetry, see introduction). Instead, we are coarse-graining to a
macrostate that provides a map from parameters to observed trajectories
that will enable us to automatically generate new parameter-trajectory
pairs that were not generated by running Eq~\ref{eq:linear_dynamics}.

We note the many-to-many mapping here means: 1) given one parameter,
different initial states will lead to different trajectories. 2)
sampling different parameters may lead to the same or similar
trajectories. We use two neural networks to learn the macroscale
relation between parameters and trajectories: one uses $\varphi_{u}$
to map the 4-d parameter matrix to a 2-d macrostate, and the other uses
$\varphi_{v}$ to map the 16-d trajectory to a 2-d macrostate (Figure~\ref{fig:linear}B), where we optimize to reduce the mutual information between the
identified macrostates in both cases (Figure~\ref{fig:linear}A).

After training, we can use the learned macrostates to design
microstates. In Figure~\ref{fig:linear}B, Given an example trajectory $T_{e}$, we can
compute its macrostate $\beta = \varphi_{v}(T_{e})$. The neural
network $\varphi_{u}^{- 1}$ samples parameters that can generate
trajectories for the example microstate (Figure~\ref{fig:linear}C). The sampled
parameters follow a conditional distribution
$P(M|\beta)$, where $M$ is the parameter
matrix. In Figure~\ref{fig:linear}C, we show how, given an anti-clockwise rotating
trajectory, the parameters sampled all lead to anti-clockwise
trajectories. By this process, we can design parameters of a system to
mimic the behavior of any example, even without needing to translate the
language describing the behavior to be human-interpretable. This ability
has broad applicability for the design and control of complex systems,
where simple mathematical descriptions have defied human scientists.
Even when we do not know or have access to how we could describe a
behavior, the neural network can still sample parameters to allow design
of new examples through self-supervised learning.

So far, we have demonstrated sampling parameters for the matrix $M$,
based on a specified macrostate (rotating anti-clockwise). We showed how
the sampled parameters allow constructing new example trajectories using
the sampled matrix $M$ in Eq~\ref{eq:linear_dynamics}
with the desired macroscale behavior. We can also sample trajectories
directly, via a sampling process where we specify the target macrostate
and then use the inverse sampling to recover trajectories. These sampled
trajectories follow the distribution of
$P\left( T \middle|\beta \right)$, where $T$
is the trajectory microstate. Figure~\ref{fig:linear}D show that the sampled
trajectories all follow the same behavior, exhibiting anti-clockwise
rotation, just as with the example trajectory. It is worth noting that
here we never had to implement Eq~\ref{eq:linear_dynamics} to generate the designed
trajectories, but they were sampled directly from identification of the
macrostate. We also did not give the neural network any concept of
``rotate'' or ``clockwise'': the neural network discovered this symmetry
on its own, as one that is relevant to how the parameters of the matrix
$M$ map to observed trajectories. This experiment gives a simple
example of how a neural network architecture like MacroNet can aid in
identifying genotype-phenotype maps, where we genotypes play the role of
parameters and phenotypes the role of trajectories.

\subsection{Simple harmonic oscillators}

Although we define macrostates on identifying symmetries underlying
general relations, time relations are still of particular interest
because of their long history in physics and their relationship to
energy. Here, we demonstrate how MacroNet can automatically identify the
symmetry of time translation invariance associated to energy, using a
simple harmonic oscillator (SHO) as a case study. The Hamiltonian of
SHOs is:

\begin{equation}
    \mathcal H = \frac{p^2}{2m}+\frac{1}{2}kx^2.\label{eq:sho}
\end{equation}

\begin{figure}[ht]
    \centering
    \includegraphics[width=6.5in]{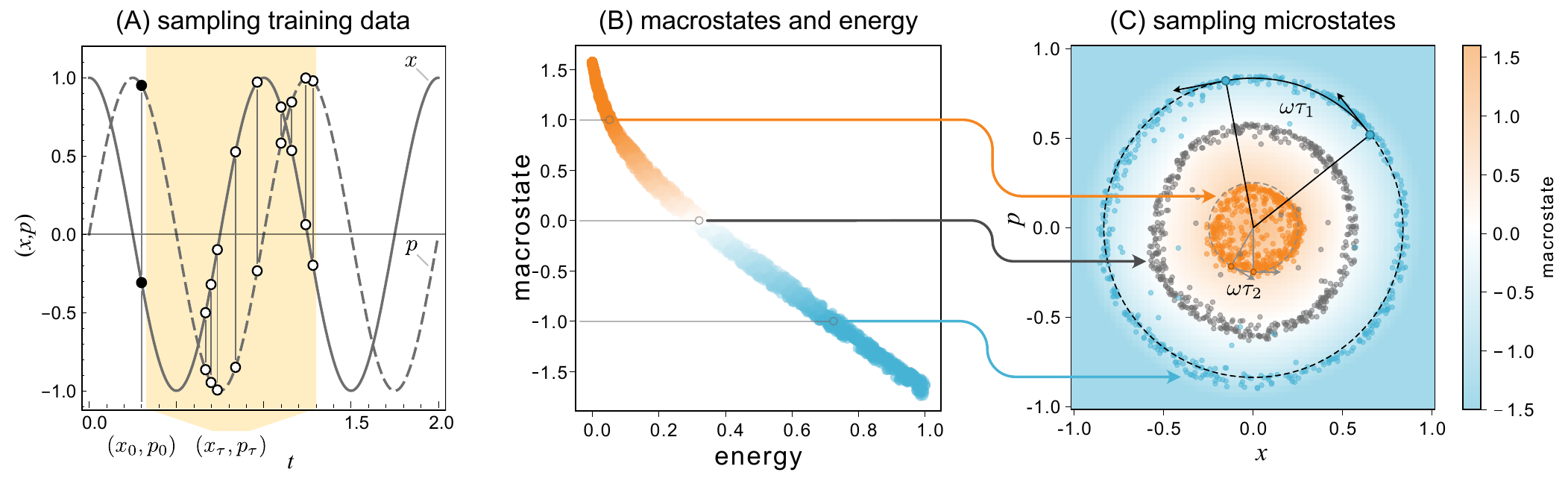}
    \caption{With a simple harmonic oscillator, we train a neural network to
find invariant quantities as a special case of macrostates. \textbf{(A)}
the $(u,v)$ pairs are sampled from simulations, where
$u = (x_{0},p_{0})$ (the black dots) and $v = (x_{\tau},p_{\tau})$.
The $\tau$ is sampled from a uniform distribution $\mathcal U(0,2\pi)$. The white dots in
the yellow region show a sampling example of $v$. Due to the
randomness of $\tau$, it is impossible for accurate prediction at
microstate. \textbf{(B)} the neural network learns energy as the
invariant quantity. The $x$-axis is the energy of microstates computed
by the physical theory of SHOs discovered by humans, and the $y$-axis
is the macrostate discovered by the neural network. They show a
monotonical relation, which implies the successful identification of
energy by the neural network. \textbf{(C)} conditional sampling
microstates from $P((x,p)|\varphi(x,p) = \alpha_{i})$, where the $\alpha_{i}$
are the given macrostates. The results approximate equal energy
surfaces, denoted by the dashed circles. Note that the noise in the
sampling is a side effect of the noisy kernel trick we use here (see
appendix). The background color also shows the learned macrostate
mapping as a field.}
    \label{fig:sho}
\end{figure}

In this experiment, we let $m = 1$ and $k = 1$ for all cases. The
micro-to-micro relation is a temporal relation, represented by pairs of
$\left( x_{0},p_{0} \right)$ and $\left( x_{\tau},p_{\tau} \right)$,
where $x_{0}$ and $p_{0}$ are the initial position and momentum and
$\tau$ is uniformly sampled time interval $(0,2 \pi)$ (see Figure~\ref{fig:sho}A). Since we
are trying to find a time invariant quantity, the mapping function
$\varphi_{u}$ and $\varphi_{v}$ should not be different. So, we
force the two neural networks $\varphi_{u}$ and $\varphi_{v}$ to
share the same weights.

Figure~\ref{fig:sho} shows our training results. When we require the neural network
to learn a one-dimensional invariant as a macrostate, the macrostate is
exactly a function of energy (Figure~\ref{fig:sho}B). Figure~\ref{fig:sho}C shows samplings from
macrostates to microstates. The same color represents microstates
sampled from the same macrostate. The sampling shows how the neural
network has identified three concentric circles, which correspond to the
equal energy surfaces of the SHOs (Figure~\ref{fig:sho}C), where the equation
$p^{2} + x^{2} = H$ represents a circle with a radius of $\sqrt{H}$.
Note that the uncertainty of $\tau$ makes it impossible to accurately
predict the future microstates. In fact, the optimal prediction at any
microstate will be zero if we optimize the MSE loss. However, using
MacroNet, we can still predict the future macrostates and sample
microstates from them. This is an example of how predictions at the
microscale can fail, and how macrostates can help solving many-to-many
mapping problems, such that predictions are still possible.

\subsection{Turing patterns}

\begin{figure}[ht]
    \centering
    \includegraphics[width=6.5in]{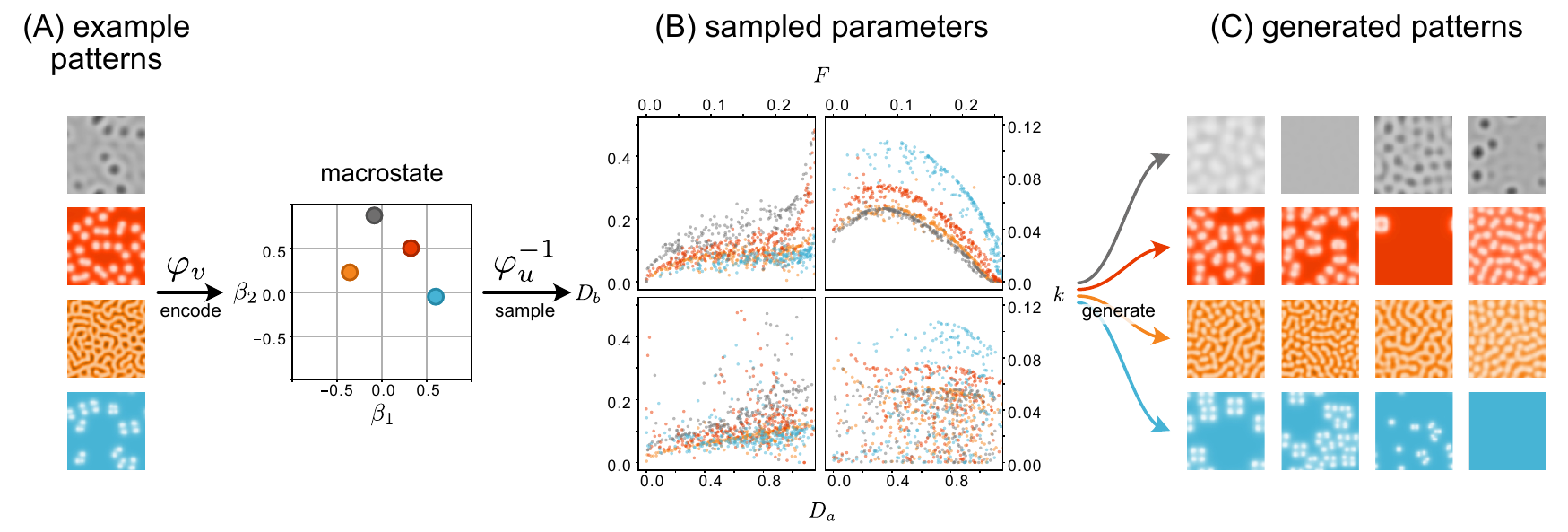}
    \caption{Experiments on Turing patterns. \textbf{(A)} By giving an example
pattern, we can compute its macrostate by $\varphi_{v}$. The patterns
are colorized for distinguishing different experiments. \textbf{(B)} Then, we can
sample an ensemble of corresponding parameters from the macrostate by
$\varphi_{u}^{- 1}$. The points with the sample color are sampled from
the same macrostate computed from the corresponding example. The dots
with black boundary are the parameters that used to generates the
example patterns. \textbf{(C)} Using the sampled parameters, we can generate
Turing patterns with sampled initial states. The generated patterns show
similar macroscopic shape as the corresponding examples.}
    \label{fig:turing}
\end{figure}

Finally, we applied the same method on a complex system: Turing
patterns. Here, we are using the Gray-Scott Model~\cite{gray1984autocatalytic}, a 2-d
space that has two kinds of components, $a$ and $b$, which might,
for example, correspond to two different kinds of chemical species. The
$a$ and $b$ are two scalar fields corresponding to concentration of
the two species. Their dynamics can be described by the differential
equation:

\begin{equation}
    \begin{split}
        \pa{a}{t} &=D_a\nabla^2 a-ab^2+F(1-a)\\
        \pa{b}{t} &= D_b \nabla^2 b +ab^2 -(F+k)b\label{eq:gary_scott}
    \end{split}
\end{equation}

where $D_{a}$, $D_{b}$, $F$ and $k$ are four positive constants
- these four parameters determine the behavior of the system. This model
can generate a set of complex patterns, see Figure~\ref{fig:turing}A. By finding
macrostates mapping patterns to parameters, we can then in turn design
related systems by specifying parameters that will yield user-specified
patterns. Here, $u$ is the parameter vector,
$u = ( D_{a},D_{b},F,k )$. And $v$ is the generated
pattern, represented by $64 \times 64$ images,
$v = \left( a^{(64 \times 64)},b^{(64 \times 64)} \right)$.

We trained the neural network to map parameters and patterns to each
other at macroscale (such that these will share the same macrostate).
Figure~\ref{fig:turing} shows the sampling based on the specified patterns. By giving
an example pattern $v$ (Figure~\ref{fig:turing}A), we can sample parameters
$u' = \varphi_{u}^{- 1}\left( \varphi_v(v),z \right)$ with the same
macrostate as $v$ (Figure~\ref{fig:turing}B). As Figure~\ref{fig:turing}C shows, the sampled rules
(set of four parameters) will generate similar patterns as the example
patterns. This experiment shows that our method can design complex
systems by sampling parameters that will generate patterns exhibiting
the same macrostate as the example behavior. That is, MacroNet can solve
the inverse problem of going from pattern to parameters.

The microstate ensembles associated with macrostates can also be
directly discovered by this approach. Figure~\ref{fig:turing}B shows the distribution
of parameters sampled from different macrostates. The sample points with
the same color are considered as equivalent to each other under the
mapping $\varphi$, which takes the microstate to a macrostate.
Parameters in the same equivalence class (sharing the same symmetry)
will therefore lead to patterns that have the same macrostates, so we
can sample any parameters along these equivalence curves and generate
Turing patterns with the user-specified behavior.

And additional feature is that observing the sampled parameters can also
tell us the importance of different parameters for specifying a target
macroscale behavior. For example, as shown in Figure~\ref{fig:turing}B, different
macrostates have similar sampling on $D_{a}$. However, on $(F,k)$,
different macrostates sample different parameters. This indicates that
$F,k$ will have stronger effect on differences in macroscale behavior
than $D_{a}$. This has implications for specifying control parameters
in designing complex systems. An example of interest is in pattern
formation in regeneration~\cite{levin2014endogenous}, where a framework like MacroNet
could identify the patterns controlling specific features of shape.

\section{Discussion}

Since Anderson published the seminal paper, \emph{More is Different}, it
has been increasingly recognized that complex systems displaying
emergent behaviors do not necessarily share the same symmetries as their
micro-rules~\cite{strogatz2022fifty}. That is, we know the mapping from a
micro-rule to a large-scale system does not preserve all the symmetries
of the micro-rule, due to symmetry breaking and perturbations from the
environment. In some sense, this is the very definition of ``emergence''.
However, we might expect some symmetries to be retained such that
micro-rules share at least a subset of their symmetries with any
macroscale emergent behavior. Indeed, this is what we see in the
experiments presented in this work. Each macrovariable can represent a
type of symmetry: for instance, the energy of a simple harmonic
oscillator represents how all states with the same energy are symmetric
in time to others with that energy. In a more complex case, the
macrostates of Turing patterns contain the information that is invariant
under the mapping from parameter to pattern, even under external
perturbations. The parameters that having the same macrostate are
symmetric to each other because they all generate the patterns with the
same macrostate. By finding the macrostates via the mutual information
shared between ensembles of microstates, we can align the symmetries
shared by the two sets of microvariables. This is a general framework
for identifying macrostates as maps conserving the symmetries of
systems: hence, while given ``more is different'' is true in most cases,
we can still find examples of macrovariables that behave as ``more is
same'' because they will retain underlying symmetries present at the
microscale.

The process of finding macrostates can be considered as a prediction
problem: that is, it is one of finding predictable variables of two
related observations. There are no such variables if two observations
have zero mutual information. Thus, if two observations have none-zero
mutual information, we can use macrovariables (ensembles of microstates)
to connect the two observations. In this way, one can consider
macrostates as the instantiated mutual information mapping observations
of one system to another (or a system to itself at a different point in
time).

Across our experiments, we showed how macrostates can emerge from
identifying predictive relations between two sets of observations. The
parameter-trajectory relation leads to the macrostate of rotation and
direction. The temporal relation between past and future leads to the
macrostate energy in the simple harmonic oscillator. In the more complex
case of Turing patterns, macrostates arise from parameter-pattern
relationships. Thus, by adopting this relationalism idea, we can
establish an approach targeting an ambitious question in the complex
systems field: is it possible find general laws of complex systems? To
address this question, one key task is to find a set of universal
macrostates that can be found in most complex systems. And hence the
laws of the universal macrostates can be considered as the general laws
of complex systems. The method proposed in this work makes an initial
step for this target -- by finding macrostates from relations, the
macrostates can be used on both sides of the relations (although they
may be interpreted differently on either side of the relation). For
instance, in the Turing pattern case, the macrostates are not only the
macrostates of patterns, but also the macrostates of parameters. For
future work, to find more universal macrostates, the framework may be
extended from second-order relationship to higher-order relationships.
Applying this method more generally to complex systems may reveal there
are indeed universal general laws, or it may reveal that no map can
apply to all systems -- that is, that the laws of complex systems are
unique to specific classes of system. In either case, the framework we
have presented here, which offers an automated means for identifying
general laws via symmetries in complex systems, offers new opportunities
for asking and answering such questions.

\section*{Acknowledgement}

We would like to acknowledge Dr. Cole Mathis, Dr. Daniel Czegel, Dr. Douglas G. Moore and Dr. Enrico Borriello from the Walker lab at Arizona State University for suggestions and discussions. We thank Prof. Yi-Zhuang You from UCSD for his valuable suggestions on machine learning, Prof. Yunbo Lu from Tongji University for his support on computational resources, and Prof. Leroy Cronin from University of Glasgow for research support. This project was supported by a grant from the John Templeton Foundation.

\section*{Author Contributions}

Y.Z. and S.I.W. developed the theory and designed the formalism. Y.Z.
implemented the MacroNet machine learning architecture and performed the
analyses. Y.Z. and S.I.W. wrote the manuscript.

\bibliographystyle{unsrt}  
\bibliography{references}

\end{document}